\def\year{2021}\relax
\documentclass[letterpaper]{article} 
\usepackage{aaai21}  
\usepackage{times}  
\usepackage{helvet} 
\usepackage{courier}  
\usepackage[hyphens]{url}  
\usepackage{graphicx} 
\usepackage{natbib}  
\usepackage{caption} 
\usepackage{xcolor}
\usepackage{amsmath,amssymb,amsfonts}
\usepackage{subcaption}
\usepackage{color, colortbl}
\definecolor{Gray}{gray}{0.9}   
\graphicspath{{./figs/}}
\title{Queue-Learning: A Reinforcement Learning Approach for\\ Providing Quality of Service}
\author {
        Majid Raeis,
        Ali Tizghadam, 
        Alberto Leon-Garcia \\
}
\affiliations {
     University of Toronto, Canada \\
    m.raeis@mail.utoronto.ca, ali.tizghadam@utoronto.ca, alberto.leongarcia@utoronto.ca
}
\begin{document}

\maketitle

\begin{abstract}
End-to-end delay is a critical attribute of quality of service (QoS) in application domains such as cloud computing and computer networks. This metric is particularly important in tandem service systems, where the end-to-end service is provided through a chain of services. Service-rate control is a common mechanism for providing QoS guarantees in service systems. In this paper, we introduce a reinforcement learning-based (RL-based) service-rate controller that provides probabilistic upper-bounds on the end-to-end delay of the system, while preventing the overuse of service resources. In order to have a general framework, we use queueing theory to model the service systems. However, we adopt an RL-based approach to avoid the limitations of queueing-theoretic methods. In particular, we use Deep Deterministic Policy Gradient (DDPG) to learn the service rates (action) as a function of the queue lengths (state) in tandem service systems. In contrast to existing RL-based methods that quantify their performance by the achieved overall reward, which could be hard to interpret or even misleading, our proposed controller provides explicit probabilistic guarantees on the end-to-end delay of the system. The evaluations are presented for a tandem queueing system with non-exponential inter-arrival and service times, the results of which validate our controller's capability in meeting QoS constraints.
\end{abstract}

\section{Introduction}\label{intro}

End-to-end delay of a service system is an important performance metric that has a major impact on the customers' satisfaction and the service providers' revenues. Therefore, providing guarantees on the end-to-end delay of a service system is of great interest to both customers and service providers. Ensuring quality of service (QoS) guarantees is often a challenging task, particularly when the service system is composed of finer-grained service components. This can be seen in many different service contexts such as cloud computing and computer networks. Specifically, with the emergence of Network Function Virtualization (NFV) in cloud environments, end-to-end service networks can be created by chaining virtual network functions (VNF) together. The goal of a cloud service provider is to efficiently manage the resources, while satisfying the QoS requirements of the service chains. This can be achieved through different control mechanisms such as vertical (horizontal) scaling of the resources, which corresponds to adding/removing CPU or memory resources of the existing VNF instances (adding/removing VNF instances) in response to the workload changes~\cite{elasticsfc}.

In order to study the problem of QoS assurance in a broader context, we take a general approach and use queueing-theoretic models for representing the service systems. Therefore, the presented results can be applied to a wide range of problems, such as QoS assurance for VNF chains. In this paper, we focus on dynamic service-rate control of the tandem systems as an effective approach for providing end-to-end delay guarantees. This is closely related to the concept of vertical auto-scaling of VNF service chains. There is a rich body of literature on the service-rate control of the service systems, the majority of which is based on queueing theoretic approaches~\cite{kumar, lee, weber}. However, most of this literature is limited to simple scenarios or unrealistic assumptions, such as exponential inter-arrival and service times. This is due to the fact that the queueing theoretic techniques become intractable as we consider larger networks under more realistic assumptions. 

\begin{figure}[t!]
\centering
\includegraphics[scale=0.4]{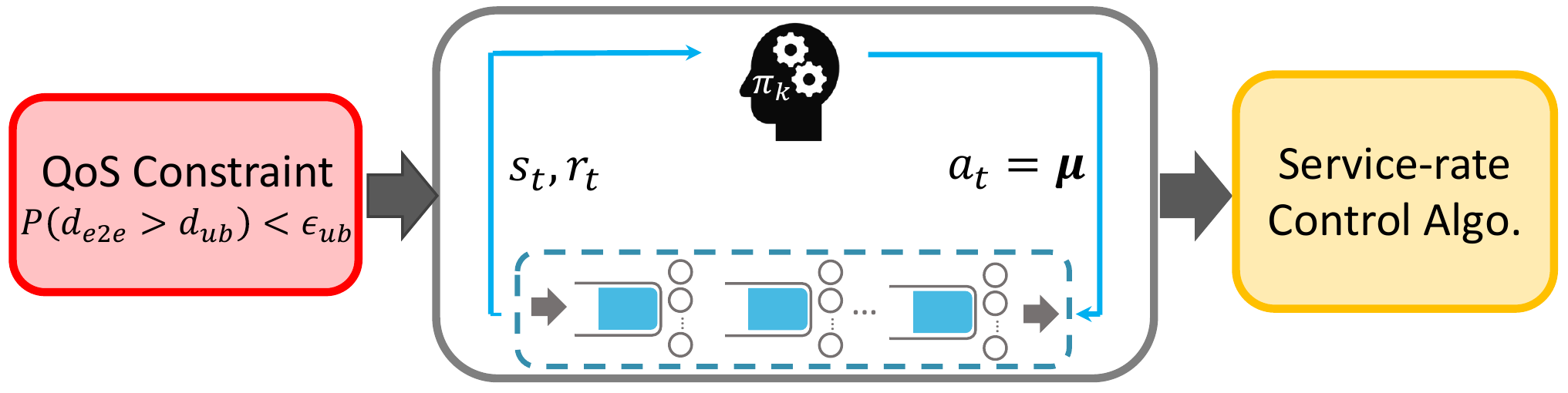}
\caption{Queue learning: RL-based framework for providing QoS guarantees}
\label{fig:qos}
\end{figure}

Considering the shortcomings of the queueing theoretic methods and the fact that the service-rate control problem involves sequential decision makings, we adopt a reinforcement learning approach, which is a natural candidate for dealing with these types of problems. In particular, we use Deep Deterministic Policy Gradient (DDPG) \cite{ddpg} algorithm to handle the continuous action space (service rates) and the large state space (queue lengths) of the problem. The contributions of this paper can be summarized as follows
\begin{itemize}

    \item We introduce an RL-based framework that takes a QoS constraint as input, and provides a dynamic service-rate control algorithm that satisfies the constraint without overuse of the service resources (Fig.~\ref{fig:qos}). This makes our method distinct from the existing service-rate control algorithms that quantify their performance by the achieved overall reward, which is highly dependent on the reward definition and might have no practical interpretations.
    
    \item The proposed controller provides explicit guarantees on the \emph{end-to-end delay} of the system. This is in contrast to existing methods that only consider some implicit notion of the system's latency such as the queue lengths.
    
    \item  Our RL-based controller provides \emph{probabilistic upper-bounds} on the end-to-end delay of the system. This is an important contribution of this paper since ensuring probabilistic upper-bounds on a particular performance metric, such as the end-to-end delay of the system, is a much more challenging task compared to the common practice control methods that only consider average performance metrics such as the mean delay.
    
    
\end{itemize}

\subsection{Related Work}
Service system control problem has been studied using various techniques. The majority of these studies are based on queueing theoretic methods, which are often limited to simple and unrealistic scenarios due to mathematical intractability of the problem. The common approach used in this class of studies is to minimize a cost function that consists of different terms such as the customer holding cost and the server operating cost~\cite{kumar, lee}. These methods do not directly consider the delay of the system in their problem formulation and therefore, cannot provide any delay guarantees. Using a constraint programming approach, Terekhov et al.~\shortcite{terekhov} study the control problem of finding the smallest-cost combination of workers in a facility, while satisfying an upper-bound on the expected customer waiting time. Although this work provides a guarantee on the delay of the system, the authors only consider the average waiting time of the system. Another approach that has recently emerged as a popular control method for complex service systems is reinforcement learning~\cite{liu, raeis}. The authors in these works study queueing control problems such as server allocation, routing~\cite{liu} and admission control~\cite{raeis}.  

Similar control problems have been studied in different areas such as cloud computing and computer networks. VNF auto-scaling is one such example, which aims to provide QoS guarantees for service function chains~\cite{elasticsfc,dynamic,auto}. However, the proposed algorithms in this area are often based on heuristic methods and do not provide much insight into the behaviour of the system or the control mechanism. 

\section{Problem Model and Formulation}\label{sys_model}

\begin{figure}[t!]
\centering
\includegraphics[scale=0.33]{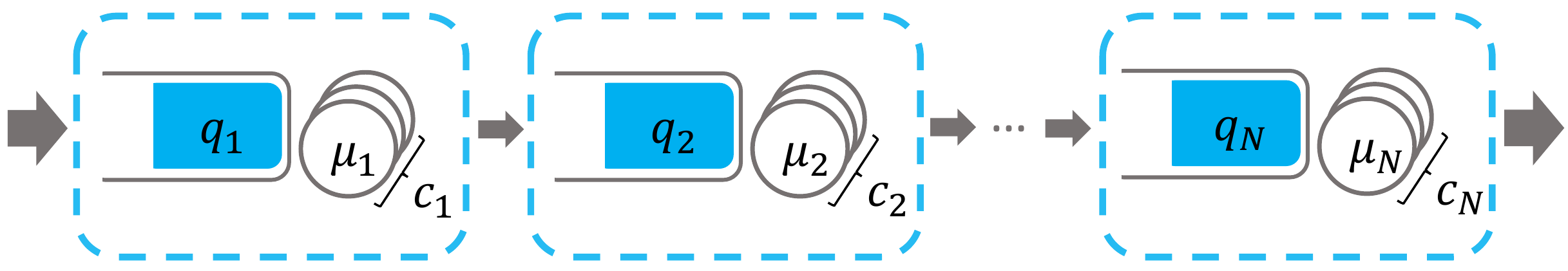}
\caption{System model: a tandem network of multi-server queueing systems, where the service rates $\mu_n$, $1\leq n \leq N$, can be dynamically adjusted by the controller to satisfy the QoS constraint.}
\label{fig:tandem}
\end{figure}

In this paper, we consider multi-server queueing systems, with First Come First Serve (FCFS) service discipline, as the building blocks of the tandem service networks that we study (Fig.\ref{fig:tandem}). In a tandem topology, a customer must go through all the stages to receive the end-to-end service. We do not assume specific distributions for the service times or the inter-arrival times and therefore, these processes can have arbitrary stationary distributions.
We study QoS assurance problem for a tandem network of $N$ queueing systems, where system $n$, $1\leq n \leq N$, is a multi-server queueing system with $c_n$ homogeneous servers having service rates $\mu_n$. Moreover, we consider a service rate controller that chooses a set of service rates for the constituent queueing systems  every $T$ seconds. Therefore, the controller takes an action at the beginning of each time slot\footnote{Throughout the paper, we use the terms \emph{time slot} and \emph{time step} interchangeably.} and the service rates are fixed during the time slot. We denote by $q_n$ the queue length of the $n$th queueing system at the begining of each time slot. The service rate controller takes actions based on the queue length information of all the constituent systems, i.e., $(q_1,q_2,\cdots,q_N)$. Moreover, we denote the end-to-end delay of the system at an arbitrary time slot by $d$.

As mentioned earlier, the reason for considering the service rate controller is to provide QoS guarantees on the end-to-end delay of the system. More specifically, our goal in designing the controller is to provide a probabilistic upper-bound on the end-to-end delay of the customers, i.e., $P(d>d_{ub})\leq \epsilon_{ub}$, where $d_{ub}$ denotes the delay upper-bound and $\epsilon_{ub}$ is the violation probability of the constraint. Since there is an intrinsic trade-off between the system's service capacity and its end-to-end delay, satisfying the QoS constraint might be achieved at the cost of inefficient use of the service resources. Therefore, the amount of service resources that are used by the controller is an important factor that needs to be considered in our algorithm design. 

\section{Service Rate Control as a Reinforcement Learning Problem}\label{AC_RL}
In this section, we start with a brief review of some basic concepts and algorithms in reinforcement learning. Then, we formulate our problem in the RL framework. 
\subsection{Background on Reinforcement Learning}

The basic elements of a reinforcement learning problem are the \emph{agent} and the \emph{environment}, which have iterative interactions with each other. The environment is modeled by a Markov Decision Process (MDP), which is specified by $<\mathcal{S},\mathcal{A},\mathcal{P},\mathcal{R}>$, with state space $\mathcal{S}$, action space $\mathcal{A}$, state transition probability matrix $\mathcal{P}$ and reward function $\mathcal{R}$. At each time step $t$, the agent observes a state $s_t \in \mathcal{S}$, takes action $a_t \in \mathcal{A}$, transits to state $s_{t+1} \in \mathcal{S}$ and receives a reward of $r_t=\mathcal{R}(s_t,a_t,s_{t+1})$. The agent's actions are defined by its policy $\pi$, where $\pi(a|s)$ is the  probability of taking action $a$ in state~$s$. The total discounted reward from time step $t$ onwards is called the return, which is calculated as $r_t^\gamma = \sum_{k=t}^\infty \gamma^{k-t} r_k$, where $0 < \gamma < 1$. The goal of the agent is to find a policy that maximizes the average return from the start state, i.e., $\max_{\pi}$ $J(\pi)=\mathbb{E}[r_1^\gamma|\pi]$.
Let us define the (improper) discounted state distribution as $\rho^{\pi}(s')=\int_\mathcal{S}{\sum_{t=1}^\infty \gamma^{t-1} p(s)p(s\to s',t,\pi)\mathrm{d}s}$, where $p_1(s)$ and $p(s\to s',t,\pi)$ denote the density at the starting state and the density at state $s'$ after transitioning for $t$ steps from state $s$, respectively. Now, we can summarize the performance objective as $J(\pi)= \mathbb{E}_{s \sim \rho^\pi, a\sim \pi}[r(s,a)]$.

\subsubsection{Stochastic Policy Gradient} Policy gradient is the most popular class of continuous action reinforcement learning algorithms. In these algorithms, the policy function is parametrized by $\theta$ and is denoted by $\pi_\theta(a|s)$. The parameters $\theta$ are adjusted in the direction of $\nabla_\theta J(\pi_\theta)$ to optimize the performance function. Based on the policy gradient theorem~\cite{policy_grad},  we have
\begin{align}\label{eq:st_pol_grad}
    \nabla_\theta J(\pi_\theta) = \mathbb{E}_{s \sim \rho^\pi, a\sim \pi_\theta}[\nabla_\theta \log{\pi_\theta (a|s)}Q^\pi(s,a)],
\end{align}
where $Q^\pi(s,a)$ denotes the action-value function defined as $Q^\pi(s,a)=\mathbb{E}[r_1^\gamma|S_1=s, A_1=a;\pi]$. To calculate the gradient in Eq.~(\ref{eq:st_pol_grad}), policy gradient methods often use a sample based estimate of the expectation. However, an important challenge facing these methods is the estimation of the unknown action-value function $Q^\pi(s,a)$.
\subsubsection{Stochastic Actor-Critic } This family of policy gradient algorithms use an \emph{actor-critic} architecture to address the above mentioned issue~\cite{policy_grad, natural_ac}. In these algorithms, an actor adjusts the parameters $\theta$ of the policy, while a critic estimates an action-value function $Q^w(s,a)$ with parameters $w$, which is used instead of the unknown true action-value function in Eq.~(\ref{eq:st_pol_grad}).
\subsubsection{Deterministic Policy 
Gradient (DPG) } In contrast to the policy gradient methods, DPG~\cite{dpg} uses  a deterministic policy $\mu_\theta: \mathcal{S}\to\mathcal{A}$. Since the randomness only comes from the states, the performance objective can be written as $J(\mu_\theta)= \mathbb{E}_{s \sim \rho^\mu}[r(s, \mu_\theta(s))]$. Moreover, using the deterministic policy gradient theorem, the gradient of the performance objective can be obtained as 
\begin{align}\label{eq:det_pol_grad}
    \nabla_\theta J(\mu_\theta) = \mathbb{E}_{s \sim \rho^\mu} \big[\nabla_\theta \mu_\theta(s) \nabla_a Q^\mu (s,a)|_{a=\mu_\theta(s)}\big].
\end{align}
The same notions of actor and critic can be used here. Comparing Eqs.~(\ref{eq:st_pol_grad}) and (\ref{eq:det_pol_grad}), we can observe that the expectation in the deterministic case is taken only with respective to the state space, which makes DPG more sample efficient than the stochastic policy gradient, especially for problems with large action spaces.
\subsubsection{Deep Deterministic Policy 
Gradient (DDPG) }
This algorithm generalizes DPG to large state spaces by using neural network functions, which approximate the action-value function~\cite{ddpg}. While most optimization algorithms require i.i.d samples, the experience samples are generated sequentially in an environment. To address this issue, DDPG uses a \emph{replay buffer}. Specifically, the experiences are stored in the replay buffer after each interaction, while a minibatch that is uniformly sampled from the buffer is used for updating the actor and critic networks. Moreover, to deal with the instability issue that is caused in the update process of the Q-function, DDPG uses the concept of \emph{target} networks, which is modified for the actor-critic architecture. The target networks $Q'$ and $\mu'$, with parameters $w'$ and $\theta'$, are updated as $w' \leftarrow \tau w + (1-\tau) w'$ and $\theta' \leftarrow \tau \theta + (1-\tau) \theta'$, where $\tau \ll 1$. As a result, the stability of the learning process is greatly improved by forcing the targets to change slowly. 

\subsection{Problem Formulation}
Now, let us formulate the service-rate control task as a reinforcement learning problem. Our environment is a tandem queueing network as in Fig.~\ref{fig:tandem} and the agent is the controller that adjusts the service rates of the network's queueing systems. The goal is to design a controller that guarantees a probabilistic upper-bound on the end-to-end delay of the system, i.e., $P(d>d_{ub})<\epsilon_{ub}$, while minimizing the average sum of service rates per time slot.
In order to achieve this goal, our controller interacts with the environment at the beginning of each time slot. Therefore, the service rates are fixed during each time slot.

Now, let us define the components of our reinforcement learning problem as follows:
\begin{itemize}
    \item \textbf{State:} The state of the system is denoted by $s=(q_1, q_2, \cdots, q_N)$, where $q_n$ is the queue length of the $n$th queueing system at the beginning of each time step. 
    \item \textbf{Action:} The action is defined as choosing a set of service rates for the constituent queueing systems, i.e., $a~=~(\mu_1, \mu_2, \cdots, \mu_N)$. Here we consider deterministic policies and therefore, action will be a deterministic function of the state, i.e. $a=\mu_{\theta}(s)$. 
    \item \textbf{Reward:} Designing the reward function is the most challenging part of the problem. 
    The immediate reward at each time step should reflect the performance of the system, in terms of the end-to-end delay, under the taken action in that particular time slot. We pick the duration of the time slots, $T$, such that $T > d_{ub}$. Let $\mathcal{A}_t$ denote the set of arrivals at time step $t$, i.e. $[t, t+T]$, except those that arrived in $[t+T-d_{ub}, t+T]$ but did not depart until $t+T$. The reason for this exclusion is that we cannot find out if the end-to-end delay of an arrival in $\mathcal{A}^c_t$ (complement of set $\mathcal{A}_t$) will exceed $d_{ub}$ or not, by the end of that time slot. However, one should note that this portion of excluded arrivals will be negligible if $T/d_{ub}\gg1$ or $\epsilon_{ub}\ll1$. Nevertheless, we calculate the immediate reward at time step $t$, which is represented by $r_t$, only based on the arrivals in $\mathcal{A}_t$. 
    We assign sub-rewards to each arrival in $\mathcal{A}_t$ as follows: 
    \begin{equation}\label{Eq:reward}
    r'_i = \Big\{
    \begin{array}{ll}
    \beta_1\qquad  d_i<d_{ub}\\
    \beta_2\qquad  d_i>d_{ub}
    \end{array}
    , \quad i\in \mathcal{A}_t,
    \end{equation}
    where $r'_i$ and $d_i$ denote the assigned sub-reward and the end-to-end delay of the $i$th arrival in $\mathcal{A}_t$. Furthermore, we should take the cost of the chosen service rates into account. In other words, there is an intrinsic trade-off between the provided delay upper-bound and the average sum-rate of the queueing systems (resources). Let $\mu^{\text{sum}}_t$ denote the sum of the service rates at time step $t$, i.e., $\mu^{\text{sum}}_t=\sum_{n=1}^N c_n\mu_{n,t}$, where $\mu_{n,t}$ denotes the service rate of the $n$th queueing system at time step $t$. Now, we define the immediate reward at time step $t$, which is represented by $r_t$, as follows
    \begin{align}\label{Eq:reward_fun}
        r_t = r(s_t, \mu_\theta(s_t)) = \frac{\sum_{i \in \mathcal{A}_t}r'_i}{\mathbb{E}[n_a]} - \mu^{\text{sum}}_t,
    \end{align}
    where $\mathbb{E}[n_a]$ is the average number of arrivals in a given time step, which can be estimated from the previous arrivals. Now, the average reward per time step can be calculated using the defined reward function in Eq.~(\ref{Eq:reward_fun}) as follows
    \begin{align}  \label{Eq:ave_rew1}
        J(\mu_{\theta}) &= \int_\mathcal{S} p_\mu(s) \mathbb{E}[r(s, \mu_{\theta}(s))]\mathrm{d}s,
    \end{align}
    where $p_\mu(s)$ denotes the steady state distribution of the states, while following policy $\mu$. Let us define $\mathcal{B}_t$ as the set of arrivals in $\mathcal{A}_t$ for which $d_i<d_{ub}$.
    Now, by splitting $\mathcal{A}_t$ into $\mathcal{B}_t$ and $\mathcal{B}^c_t$, we have
    \begin{align*}
    &~\mathbb{E}[r(s_t, \mu_{\theta}(s_t))|s_t=s] \\&=\frac{\mathbb{E}\big[\sum_{i \in \mathcal{B}_t}\beta_1+\sum_{i \in \mathcal{B}^c_t}\beta_2\big|s\big]}{\mathbb{E}[n_a]} - \mathbb{E}[\mu^{\text{sum}}_t|s]\\
    &= \beta_1 \frac{\mathbb{E}\Big[|\mathcal{B}_t| \Big| s\Big]}{\mathbb{E}[n_a]}+\beta_2 \frac{\mathbb{E}\Big[|\mathcal{B}^c_t|\Big| s\Big]}{\mathbb{E}[n_a]} - \mathbb{E}[\mu^{\text{sum}}_t|s]\\
     &\simeq \beta_1 P(d<d_{ub}|s) + \beta_2 P(d>d_{ub}|s)- \mathbb{E}[\mu^{\text{sum}}_t|s],
    \end{align*}
    where the approximation in the last line can be obtained using the law of large numbers (LLN). 
    Therefore, using Eq.~(\ref{Eq:ave_rew1}) we have 
    \begin{align}\label{eq:ave_rew2}
        J(\mu_{\theta}) = \beta_1 P(d<d_{ub})+\beta_2 P(d>d_{ub}) - \mathbb{E}\bigg[\sum_{n=1}^N c_n\mu_n\bigg]
    \end{align}

    Now, let us choose the parameters of the reward function as follows
    \begin{equation}
        \beta_1=\epsilon_{ub} \lambda, \quad \beta_2= -(1-\epsilon_{ub})\lambda.
    \end{equation}
    Substituting the parameters and rewriting Eq.~(\ref{eq:ave_rew2}), we have
    \begin{equation}\label{eq:ave_rew3}
        J(\mu_{\theta}) = \lambda \underbrace{ \left(P(d<d_{ub})-(1-\epsilon_{ub})\right)}_{\text{QoS Constraint}} - \underbrace{\mathbb{E}\bigg[\sum_{n=1}^N c_n\mu_n\bigg]}_{\text{Average sum-rate}},
    \end{equation}
where $\lambda$ specifies the trade-off between the QoS constraint and the average sum rate. 
\end{itemize}

\section{Hyper-Parameter Selection}\label{hyper_param}
\subsection{Discussion on Trade-off Coefficient ($\lambda$)}
As we discussed in the previous section, $\lambda$ can be used to adjust the trade-off between the QoS constraint and the average sum rates. Here, we discuss the effect of this parameter on the learned policy by transforming our goal into an optimization problem.

Let us redefine our problem as learning a control policy that satisfies the QoS constraint with minimum service resources. Therefore, we can express the service-rate controller design problem as follows
\begin{align} \label{Eq:opt1}
    \max_{\theta}&\qquad-\mathbb{E}\bigg[\sum_{i=1}^N c_i\mu_i\bigg] \\
    \text{s.t.}&\qquad P(d<d_{ub})\geq 1-\epsilon_{ub} \nonumber.
\end{align}

We define the Lagrangian function associated with problem~(\ref{Eq:opt1}) as
\begin{align} \label{Eq:lag}
    L(\theta,\lambda) = -\mathbb{E}\bigg[\sum_{i=1}^N c_i\mu_i\bigg]
    + \lambda \left(P(d<d_{ub}|A)- (1-\epsilon_{ub})\right),
\end{align}
where $\lambda$ is the Lagrange multiplier associated with the QoS constraint in Eq.~(\ref{Eq:opt1}). As can be seen, the Lagrangian function is similar to the average reward obtained in Eq.~(\ref{eq:ave_rew3}). Moreover, the Lagrangian dual function is defined as $g(\lambda)~=~\max_{\theta} L(\theta, \lambda)$ and the dual problem can be written as
\begin{equation}\label{eq:dual}
    \min_{\lambda} g(\lambda), \qquad \text{s.t. }\lambda\geq 0.
\end{equation}
Therefore, $\lambda$ can be interpreted as the Lagrange multiplier of the dual problem. Moreover, maximizing the average reward $J(\mu_{\theta})$ with respect to $\theta$ will be the same as computing the Lagrangian dual function associated with problem~(\ref{Eq:opt1}). Hence, $\lambda$ can be seen as a hyper-parameter for our RL problem, where choosing the proper $\lambda$ can result in achieving the goal formulated in (\ref{Eq:opt1}). It should be noted that based on the KKT (Karush–Kuhn–Tucker) conditions, the optimal point $\lambda^*$ must satisfy $\lambda^* \left(P(d<d_{ub})- (1-\epsilon_{ub})\right)=0$. We will use these insights in the next section for better selection of the hyper-parameter $\lambda$. 

\subsection{Discussion on Time Slot Length~($T$)}
The length of the time slot is another design parameter that can affect the performance of the controller and the learned policy. In general, decreasing the time slot length provides finer-grained control over the system and can result in better optimized policies. However, choosing very small time slot lengths can cause two major practical problems. First, any controller has a limited speed due to its processing time and therefore, might not be able to interact with the environment in arbitrary short time-scales. The second and more important problem is that queueing systems do not respond to the actions instantaneously, which makes the policy learning problem even more complex. This can be especially problematic when the system is saturated~\cite{park}. As a result, the time slot length should be large enough such that the immediate reward defined in Eq.~(\ref{Eq:reward_fun}) provides a good assessment of the taken action. On the other hand, choosing very large time slot lengths can result in various issues too. As discussed earlier, larger time slot means less frequent control over the system and as a result, potentially less optimal control policies. Furthermore, if the time slot length becomes large enough such that the queueing system stabilizes after taking each action, the rewards become less dependent on the states and only assess the taken action (chosen service rates). Therefore, the learned actions become almost independent of the states, which questions the whole point of using adaptive service rate control. Another important factor to consider is the data efficiency of our algorithm. As we increase the time slot length, the controller will have less interactions with the environment and therefore, it will take a longer time for the controller to start learning useful policies.

Based on the above discussion, we should choose a time slot length that is large enough to capture the effect of the taken actions, but not too large that jeopardizes the dynamic nature of the policy. A related concept that can help us in determining a reasonable time slot length is the \emph{time constant} or the \emph{relaxation time} of the system. The relaxation time can be used as a measure of the transient behaviour of the queue, which is defined as the mean time required for any deviation of the queue length from its mean ($\bar{q}$) to return $1/e$ of its way back to $\bar{q}$~\cite{relaxation}. For a simple M/M/1 queue, the relaxation time can be approximated by $2\lambda/(\mu-\lambda)^2$~\cite{relaxation}. As can be seen, the relaxation time is a function of both the arrival rate ($\lambda$) and the service rate ($\mu$), which tremendously increases as $\rho=\lambda/\mu \to 1$, i.e. when the controller chooses a service rate close to the arrival rate. In addition to the fact that calculating the relaxation time for a complex network becomes mathematically intractable, the dependence of it on the service rates makes it dependent on the chosen actions and the policy. Therefore, we cannot use the relaxation time concept directly in our problem. However, we use a similar notion that is the core of our problem formulation, i.e., the probabilistic upper-bound on the end-to-end delay. Although we do not know the optimal policy beforehand, we know that it must guarantee the QoS constraint, i.e., $P(d>d_{ub})<\epsilon_{ub}$. Given that $\epsilon_{ub}$ is often a small probability, $d_{ub}$ can be used as an estimate of the time it takes the set of customers at a given time to be replaced with a new group of customers. Therefore, we will use $d_{ub}$ as a guideline for choosing the time slot length. We will discuss this issue further in the next section.

\section{Evaluation and Results}\label{eval}
In this section, we present our evaluations of the proposed controller under different circumstances. We first describe the experimental setup in terms of the service network topology and the technical assumptions used in the experiments. We then explain the implementation and parameter selection procedures. Finally, we discuss the experiments and results.
\subsection{Experimental Setup and Datasets}
As discussed in the previous sections, the reason why we adopt a queueing-theoretic approach for modeling our system is to provide general insights on the service-rate control problem in different applications. As a result, we consider general queueing models for the experimental evaluations. It should be noted that these models can be close approximations of the real-world service systems, since we make no assumptions on the inter-arrival or service time distributions in our design process. Moreover, our method can be used for pretraining purposes in real-world applications in which interacting with the environment and obtaining real experiences are highly expensive. 

We should again emphasize that most of the existing works on the control of queueing systems focus on congestion metrics such as the average queue length or the average delay, ignoring the tail of the distributions. Moreover, the primary goal in those studies is the optimization of a cost function, which is highly dependent on the design parameters. To the best of our knowledge, there is no existing work that is capable of providing probabilistic QoS guarantees for the network's end-to-end delay. Therefore, we focus on the performance of our proposed method and discuss the effect of different design parameters on the controller's performance.

\subsubsection{Queueing Environment}
In order to perform our experiments, we set-up our own queueing environment in Python. In this environment, multi-server queueing systems can be chained together to form a tandem service network. Moreover, the inter-arrival and service times can be generated using different distributions. Although we are interested in long-term performance metrics and there is no notion of terminal state in queueing systems, we chose an episodic environment design for some technical reasons. More specifically, we terminate each episode if the end-to-end delay of the network exceeds $Delay\_Max$ or the number of steps reaches $Max\_Step$. The reason why we consider $Delay\_Max$ is to achieve more efficient learning by avoiding saturated situations. This is particularly important in the beginning episodes, where random exploration can easily put the system in saturated states, in which the controller is not able to fix the situation since it has not been trained enough.  In order to make sure that the controller experiences different states for enough number of times and does not get stuck in some particular states, we terminate each episode after $Max\_Step$ steps and reset the environment. The controller can interact with the environment every $T$ seconds, and receives the next state and the corresponding reward, which is defined based on Eq.~(\ref{Eq:reward_fun}). 
\subsection{Implementation Parameters and Challenges}
The algorithm and environment are both implemented in Python, where we have used PyTorch for DDPG implementation. Our experiments were conducted on a server with one Intel Xeon E5-2640v4 CPU, 128GB memory and a Tesla P100 12GB GPU. The actor and critic networks consist of two hidden layers, each having 64 neurons. We use RELU activation function for the hidden layers, Tanh activation for the actor's output layer and linear activation function for the critic's output layer. The learning rates in the actor and critic networks are $10^{-4}$ and $10^{-3}$, respectively. We choose a batch size of $128$ and set $\gamma$ and $\tau$ to $0.99$ and $10^{-2}$, respectively. For the exploration noise, we add Ornstein-Uhlenbeck process to our actor policy~\cite{ddpg}, with its parameters set to $\mu=0$, $\theta=0.15$ and $\sigma$ decaying from $0.5$ to $0.005$. 

An implementation challenge that requires more discussion is the range of state values (queue lengths) and the actions (service rates) that should be used in the training. Specifically, we truncate $q_i$s by $q^{max}$, and limit $\mu_i$ between $\mu^{max}_i$ and $\mu^{min}_i$. The reason for considering $q^{max}$ is that when queue lengths become too large, which happens when the system is congested, the exact values of the queue length become less important to the controller. Therefore, we truncate the state components by $q^{max}=1024$ to make the state space smaller and the learning process more efficient, without having much effect on the problem's generality. It should be mentioned that we only truncate the states (observations) and not the actual queue lengths of the environment. On the other hand, the service rates should be chosen such that the system stays stable and do not get congested. Therefore, we choose $\mu^{min}_i$ such that $\rho_i<1$, where $\rho_i=\lambda_i/\mu_i$ is the traffic intensity of the $i$th queue. $\mu^{max}_i$ represents the maximum service rate of the $i$th system, which can be chosen based on the service system's available resources.

\subsection{Results and Discussion}

\begin{figure}[t!]
\centering
\includegraphics[scale=0.4]{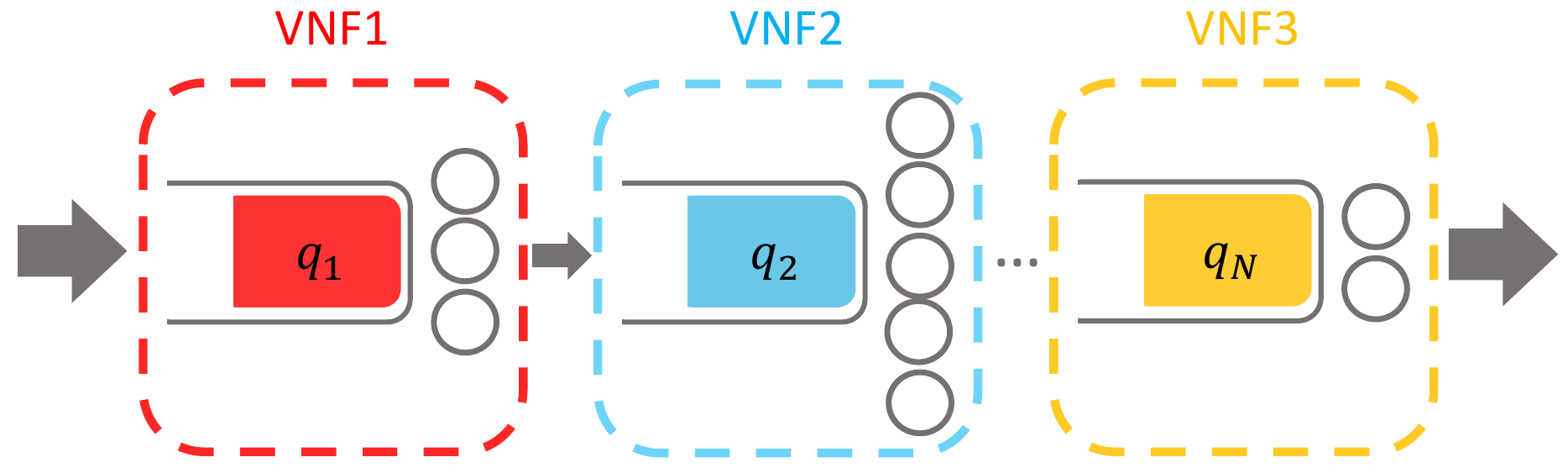}
\caption{VNF chain modeled as a tandem queueing network}
\label{fig:chain}
\end{figure}

Consider a service chain consisting of three VNFs, each modeled as a multi-server queueing system as in Fig.~\ref{fig:chain} ($[c_1,c_2,c_3]=[3,5,2]$). For this experiment, we choose Gamma distributed inter-arrival and service times. Throughout this  section, time is normalized such that the average number of arrivals per unit of time (i.e., arrival rate) equals $\lambda_a=0.95$. Moreover, the Squared Coefficient of Variation (SCV) for the inter-arrival and the service times are equal to $0.7$ and $0.8$, respectively.  We should emphasize that our design does not rely on any particular inter-arrival or service time distribution. The reason for choosing Gamma distribution is to both show our method's performance for non-exponential cases, which is often avoided for mathematical tractability, and also because this is an appropriate model for task completion times~\cite{simulation}. Our goal is to adjust the service rates ($\mu=1/Ave\_ServiceTime$) dynamically to guarantee a probabilistic delay upper-bound of $d_{ub}=10$ with violation probability $\epsilon_{ub}=0.1$, i.e., $P(d>d_{ub})\leq\epsilon_{ub}$, while minimizing the average sum service rate per decision time step. In this section, we first discuss how the reward function's hyper-parameter $\lambda$ should be chosen. Then, we present our results on the convergence of our algorithm and the effect of the time slot length on the training phase. Finally, we will compare the performance of the learned algorithms for different time slot lengths.


\subsubsection{Tuning $\lambda$ in Reward Function} 
\begin{figure}[!t] 
\centering
\subcaptionbox{}{\includegraphics[width=.48\columnwidth]{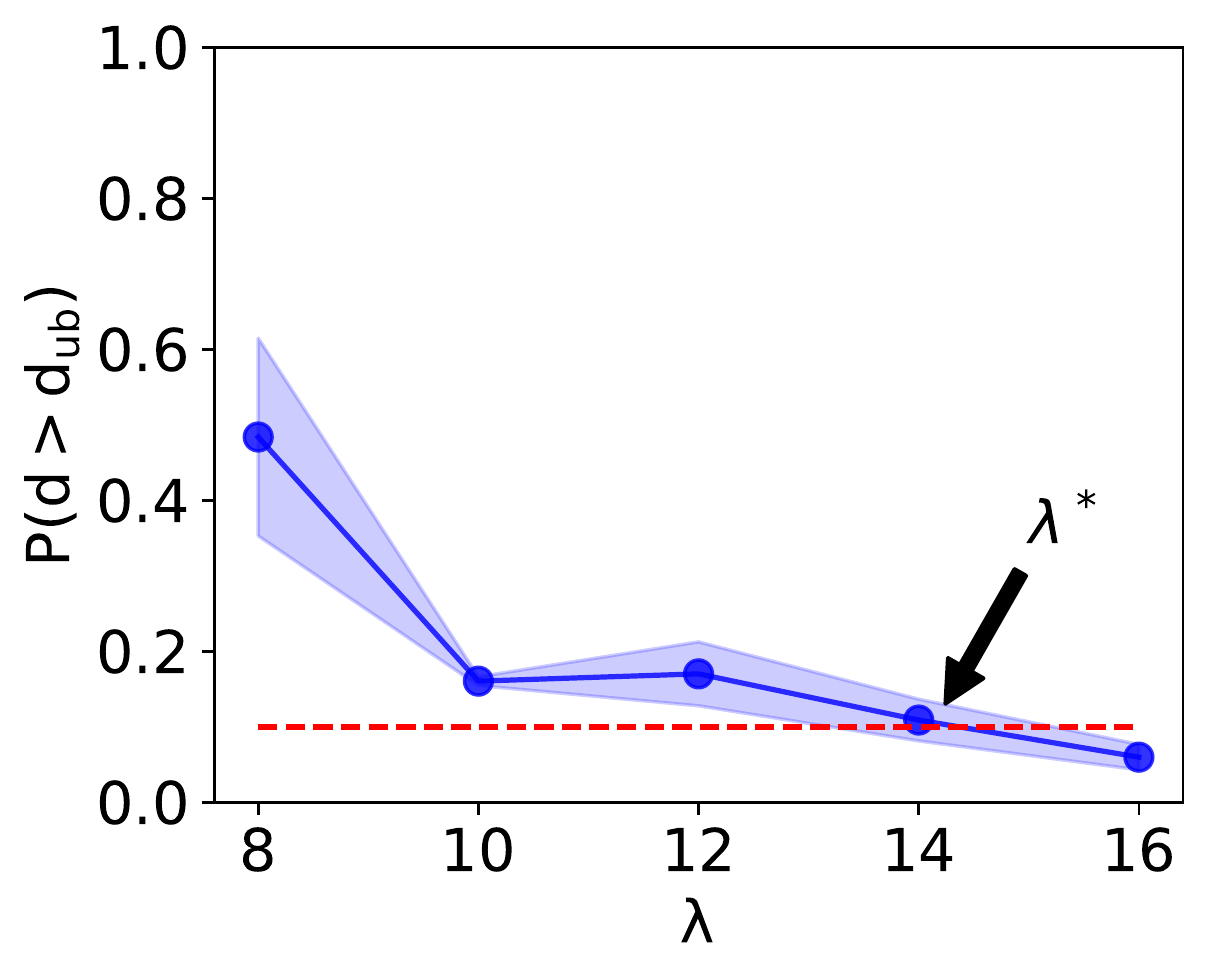}}
\subcaptionbox{}{\includegraphics[width=.48\columnwidth]{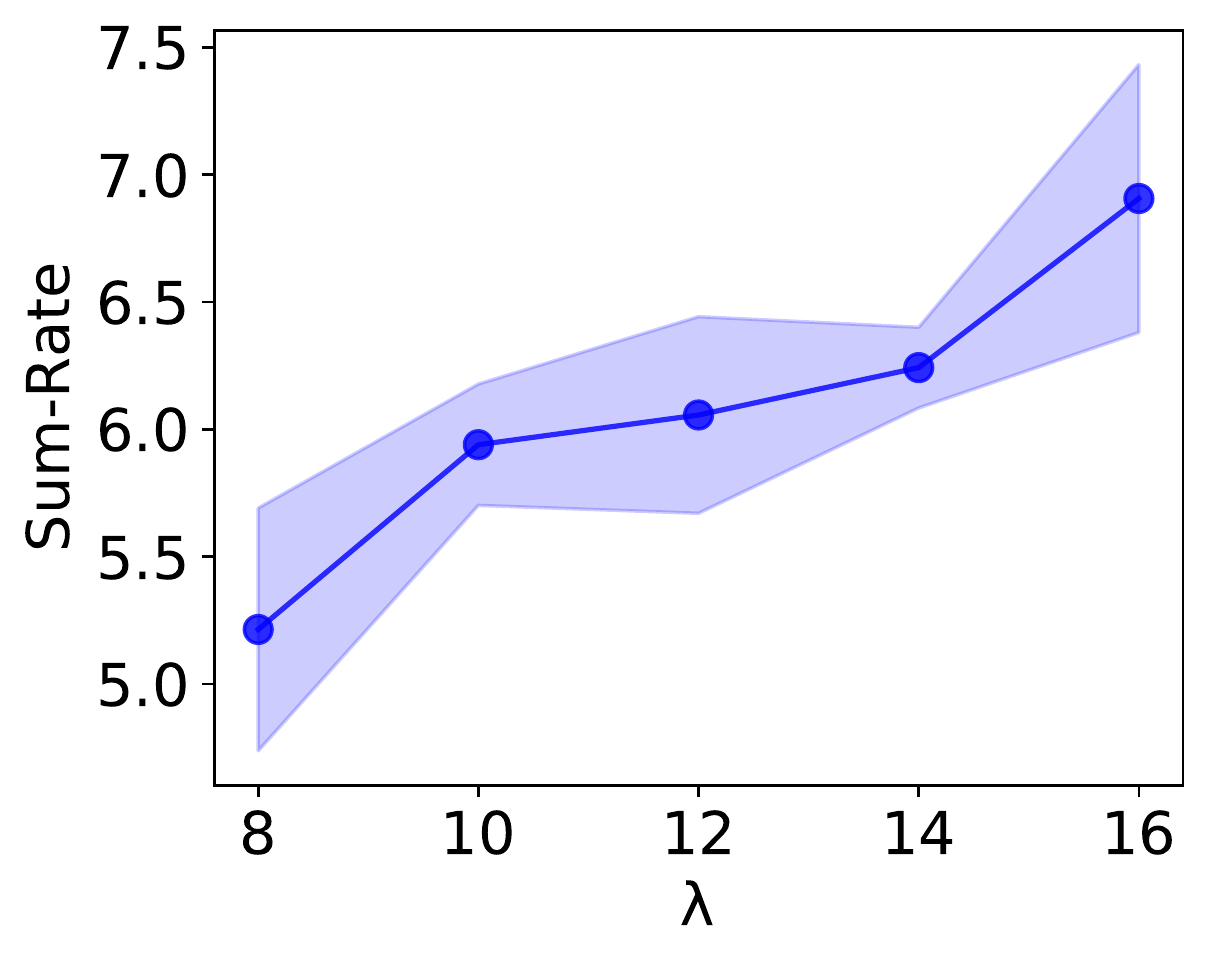}}
\caption{Performance of the controller for different values of $\lambda$: a) QoS violation probability b) Average sum service rate ($\mathbb{E}[\sum_{n=1}^N c_n\mu_n]$)}
\label{fig:Prob_violation_lambda}
\vspace{-0.5cm}
\end{figure}

As discussed earlier, we can use hyper-parameter $\lambda$ to adjust the trade-off between the QoS constraint and the consumed resources (sum-rate).
Fig.~\ref{fig:Prob_violation_lambda} shows the achieved violation probability $P(d>d_{ub})$ and the average sum service-rate ($\mathbb{E}[\sum_{n=1}^N c_n\mu_n]$), for $\lambda \in \{8,10,12,14,16\}$ and time slot length $T=30$, after training. As can be observed, increasing $\lambda$ results in the decrease of constraint violation probability and the increase of average sum-rate. Specifically, small values of $\lambda$ correspond to the case that the constraint has been removed from the reward function in Eq.(\ref{eq:ave_rew3}), which results in large violation of the QoS. On the other hand, a large $\lambda$ corresponds to the case that the sum-rate is removed from the reward function and therefore, the goal of the controller is simplified to minimization of the probability of QoS violation, which can result in tremendous overuse of the resources. 

In this experiment, the best performance is achieved for $\lambda^*\simeq14$, since it results in the minimum sum-rate, while satisfying the constraint ($P(d>10)\leq 0.1$). This can be also verified by our earlier discussion on the optimal $\lambda^*$, where the KKT (Karush–Kuhn–Tucker) condition requires that $\lambda^*\left(P(d<d_{ub})- (1-\epsilon_{ub})\right)=0$.
Therefore, in order to adjust $\lambda$, we train the controller for a range of $\lambda$ values and pick the one or which $P(d>d_{ub})$ is the closest to $\epsilon_{ub}$. 

\begin{figure*}[!t] 
\centering
\subcaptionbox{}{\includegraphics[width=.29\linewidth, height=3.1cm]{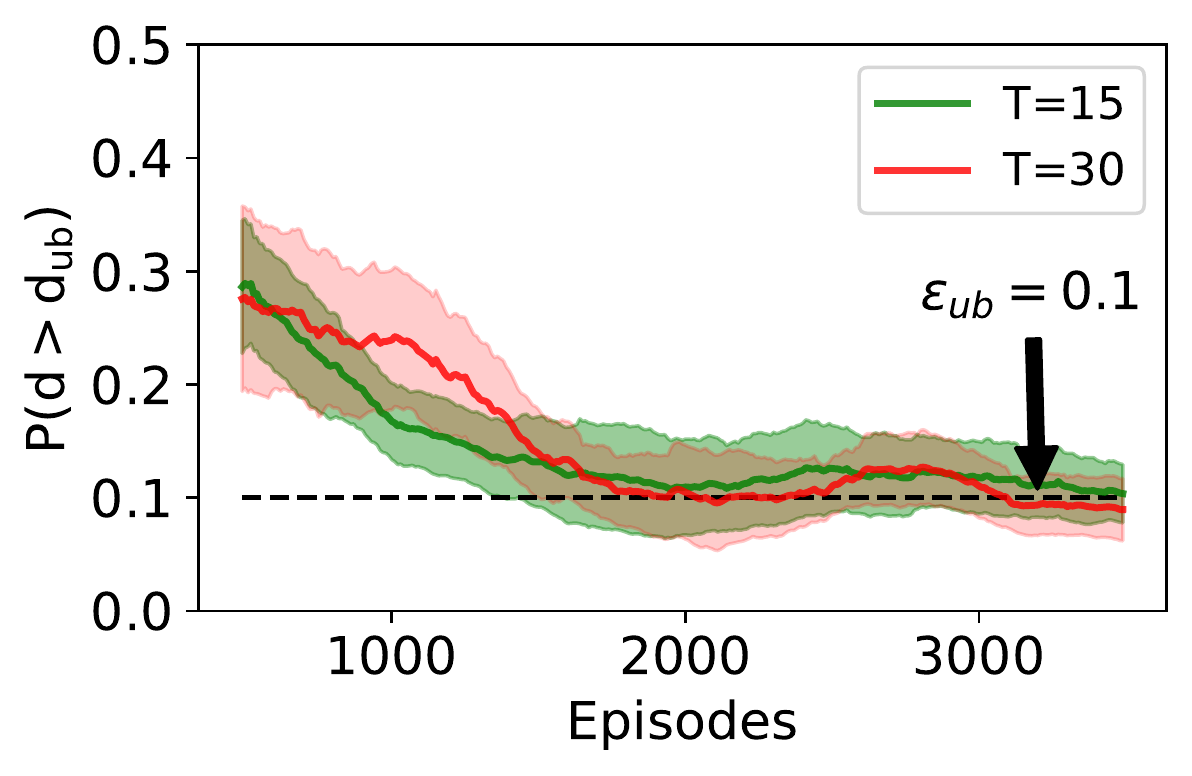}}
\subcaptionbox{}{\includegraphics[width=.29\linewidth, height=3cm]{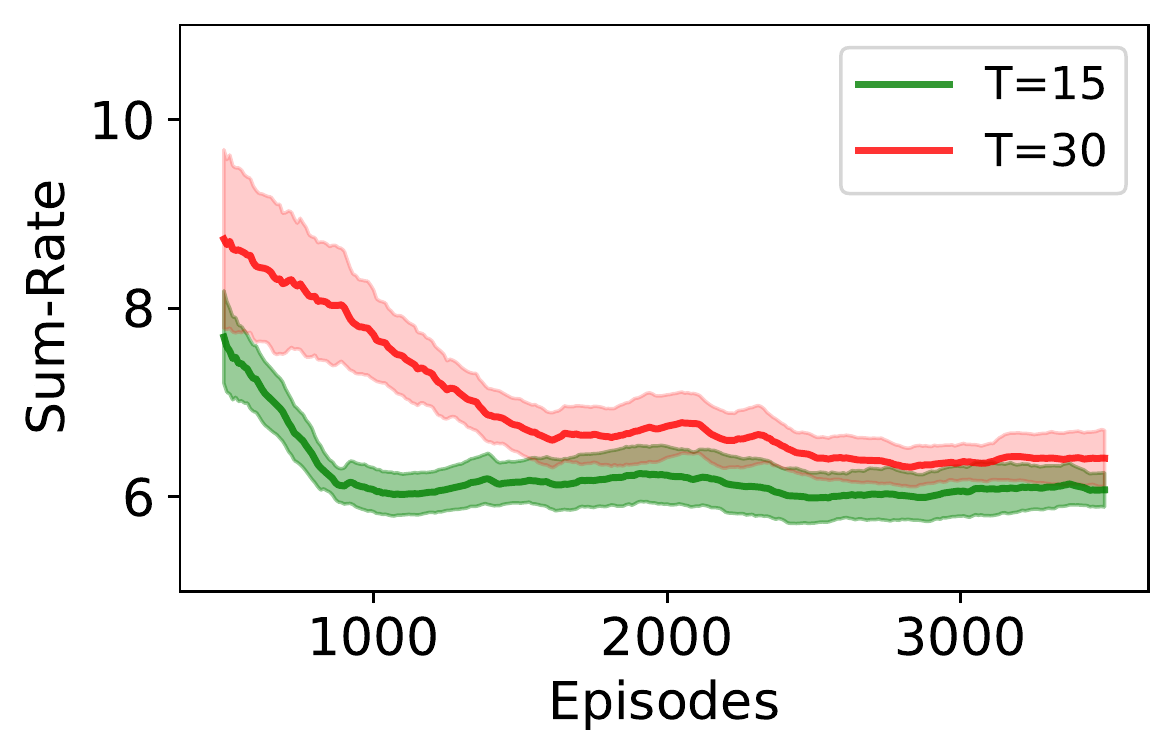}}
\subcaptionbox{}{\includegraphics[width=.29\linewidth, height=3cm]{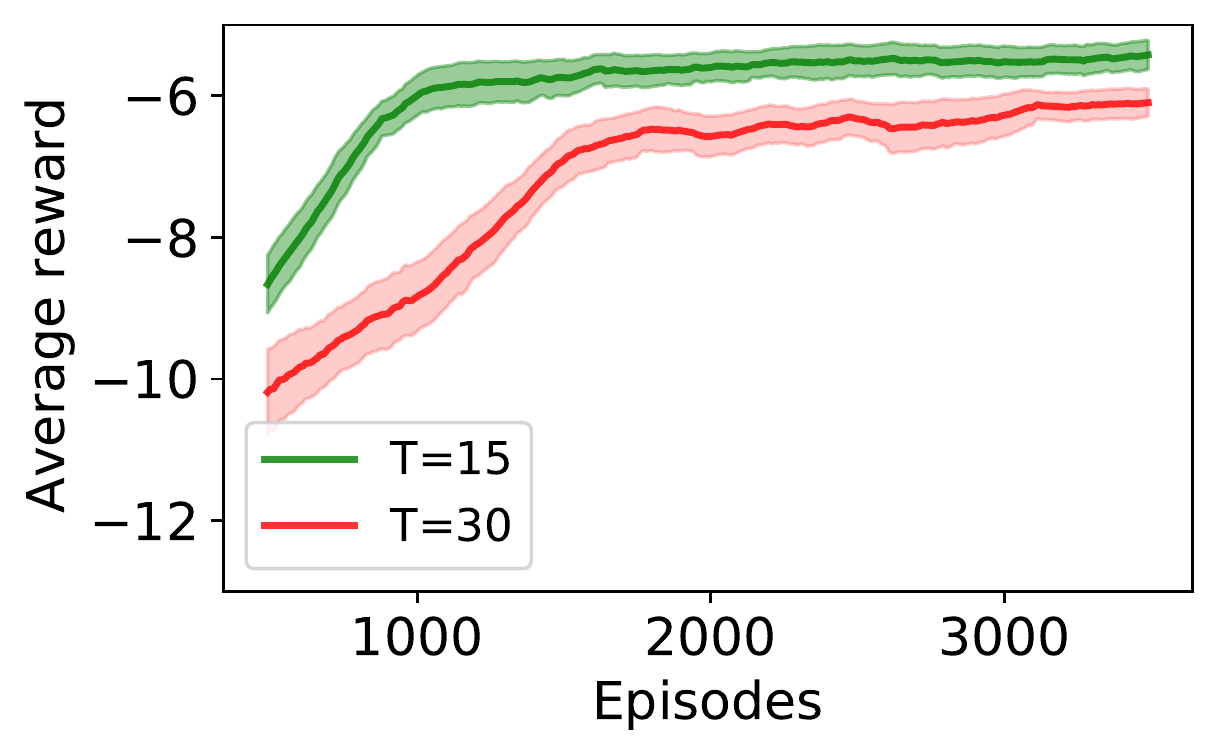}}
\hfill
\subcaptionbox{}{\includegraphics[width=.29\linewidth, height=3.1cm]{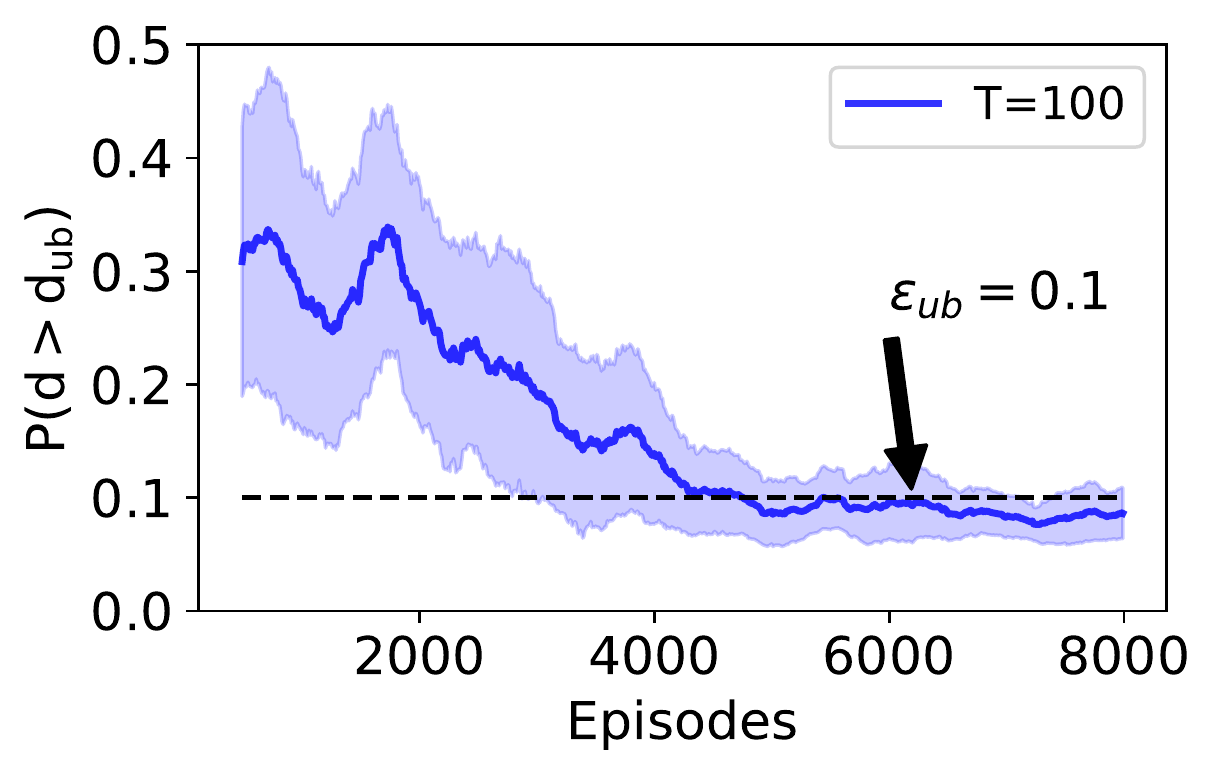}}
\subcaptionbox{}{\includegraphics[width=.29\linewidth, height=3cm]{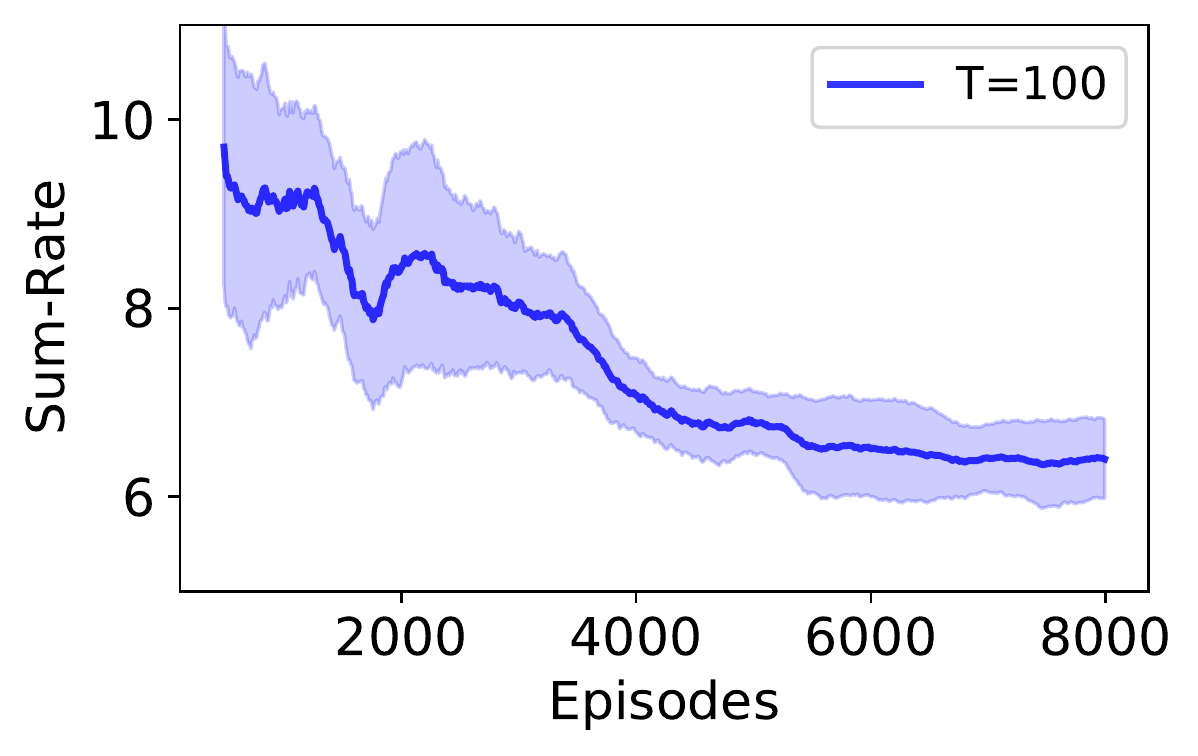}}
\subcaptionbox{}{\includegraphics[width=.29\linewidth, height=3cm]{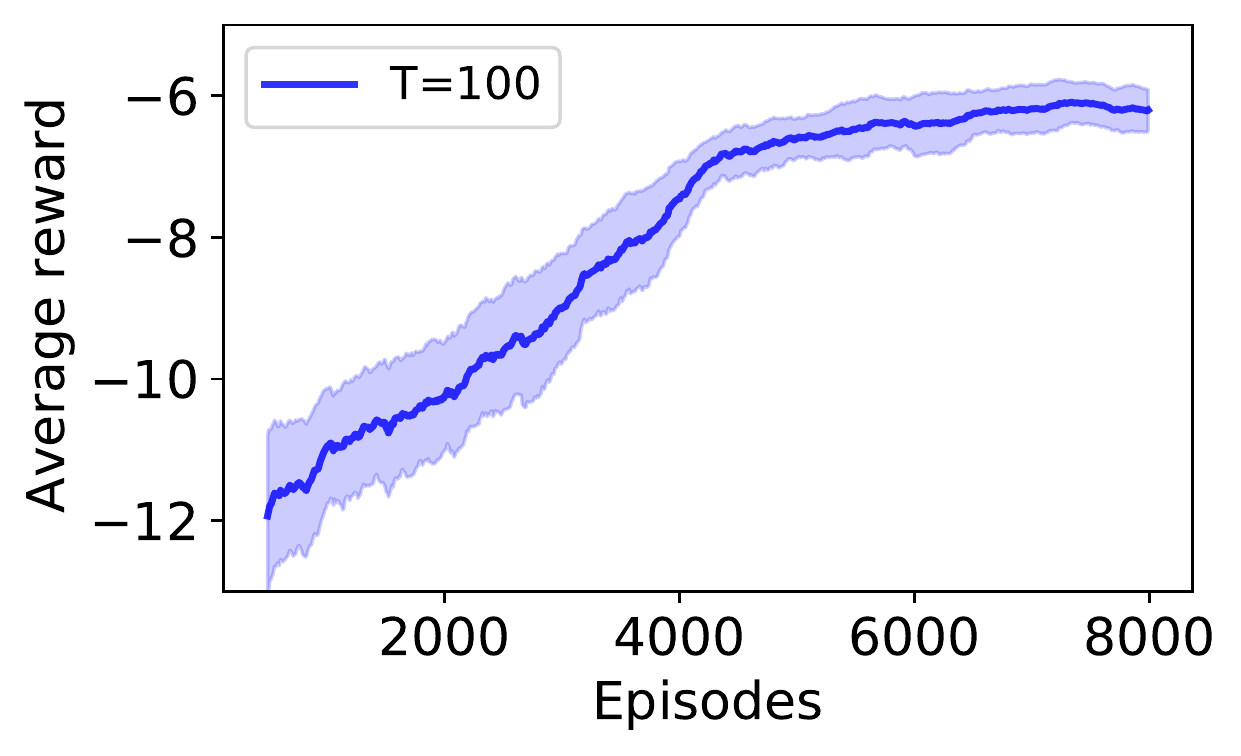}}
\caption{Impact of the time slot length on the convergence of QoS violation probability ($P(d>d_{ub})$), average sum service-rate ($\mathbb{E}\big[\sum_{i=1}^N c_i\mu_i\big]$), and average reward per time step   }
\label{fig:convergence}
\end{figure*}

\begin{figure}[t!]
\centering
\includegraphics[scale=0.5]{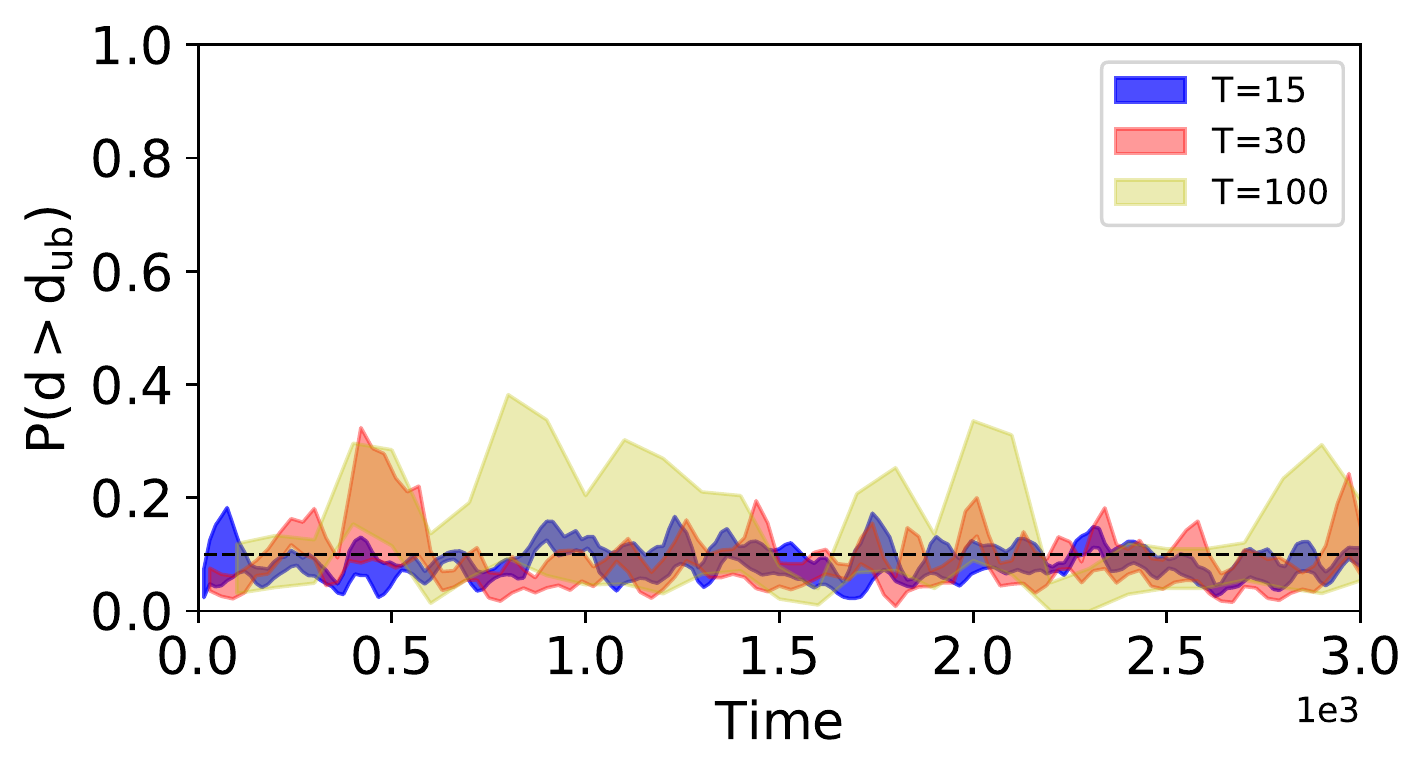}
\caption{Short-term violation of QoS constraint (Comparison of the trained controllers with different time slot lengths)}
\label{fig:test}
\end{figure}

\subsubsection{Training} Fig.~\ref{fig:convergence} shows the convergence of the violation probability ($P(d>d_{ub})$),  the average sum-rate, and the average reward per time step for three controllers with different time slot lengths of $T=15, 30$ and $100$, as a function of the number of training episodes. For each time slot length, $\lambda^*$ has been obtained using the same procedure discussed above, and the controllers have been trained with 4 different initial seeds. The dark curves and the pale regions in each figure show the average and the standard error bands, respectively.
Since the time slot lengths are different, we adjust the number of steps per episode accordingly to ensure that each episode has the same interval length of 2000 (time is normalized). As a result, the controllers with smaller time slot lengths will have more interactions with the environment during each episode and therefore, are updated more often. As can be observed from Figs.~\ref{fig:convergence} (a) and (d), increasing the time slot lengths  results in larger variations in the beginning training episodes. Moreover, the controllers with larger time slot lengths have slower convergence (the controller with $T=100$ is shown separately because of its slower convergence). It should be noted that all of the controllers roughly see the same number of departures in an episode, since the episodes are adjusted to have the same interval duration. Therefore, increasing the time slot length makes the algorithm less data efficient. Although controllers with larger time slot lengths have the advantage of receiving more meaningful rewards, since the effect of the taken action has been assessed for a longer period of time, the results suggest that time slot lengths in the order of $d_{ub}=10$ result in better performances. We can justify this using our previous discussion on the system's time constant. As mentioned earlier, $d_{ub}$ plays the same role as the time constant of the system, when $\epsilon_{ub}\ll1$. Figs.\ref{fig:convergence} (b) and (e) show that the controller with smaller time slot length is able to consume less resources (sum-rate), while satisfying the QoS constraint. Similarly, we can observe from Figs.\ref{fig:convergence} (c) and (f) that the controller with smaller time slot length can achieve a better average reward after training.

\subsubsection{Test} Now, let us compare the performance of the  trained controllers on the test data, which is generated using our environment with the same parameters discussed earlier in this section. Fig.~\ref{fig:test} plots the short-term fluctuations of the QoS violation probabilities for controllers with different time slot lengths. These violation probabilities have been calculated for short time intervals of length 1000 and can be used as short-term performance measures. Each controller has been tested with 5 different sample arrivals. Similar to the previous figures, the pale regions show the standard error bands around the means. As can be observed, the controller with the smaller time slot length ($T=15$) is more stable and does a better job of providing QoS, by controlling the violation probability at the fixed level of 10\% with small variations. On the other hand, increasing the time slot length results in larger short-term fluctuations of the QoS, because of having less control over the system.

\section{Conclusions}\label{con}
    This paper studies a reinforcement learning-based service rate control algorithm for providing QoS in tandem queueing networks. The proposed method is capable of guaranteeing probabilistic upper-bounds on the end-to-end delay of the system, only using the queue length information of the network and without any knowledge of the system model. Since a general queueing model has been used in this study, our method can provide insights into various application, such as VNF chain auto-scaling in the cloud computing context. For the future work, it would be interesting to extend this method to more complex network topologies, where a centralized controller might not be a practical solution.
\bibliography{reference}

\end{document}